\documentclass{article}
\usepackage{ijcai22}

\usepackage{times}
\usepackage{soul}
\usepackage{url}
\usepackage[hidelinks]{hyperref}
\usepackage[utf8]{inputenc}
\usepackage[small]{caption}
\usepackage{graphicx}
\usepackage{amsmath}
\usepackage{amsthm}
\usepackage{booktabs}
\usepackage{algorithm}
\usepackage{algorithmic}
\usepackage{color} 
\usepackage{bbding}
\usepackage{amssymb}
\definecolor{codeblue}{rgb}{0.25,0.5,0.5}
\definecolor{codekw}{rgb}{0.85, 0.18, 0.50}

\newcommand{\floor}[1]{\lfloor #1 \rfloor}

\title{RePre: Improving Self-Supervised Vision Transformer with \\ Reconstructive Pre-training}

\author{
Luya Wang$^1$
\and
Feng Liang$^2$\and 
Yangguang Li$^3$\and 
Honggang Zhang$^1$\and
Wanli Ouyang$^4$\and
Jing Shao$^3$
\affiliations
$^1$Beijing University of Posts and Telecommunications\\
$^2$University of Texas at Austin\\
$^3$SenseTime\\
$^4$The University of Sydney
\emails
wangluya@bupt.edu.cn,
jeffliang@utexas.edu, 
liyangguang@sensetime.com
}

\begin{document}

\maketitle

\begin{abstract}
Recently, self-supervised vision transformers have attracted unprecedented attention for their impressive representation learning ability. 
However, the dominant method, contrastive learning, mainly relies on an instance discrimination pretext task, which learns a global understanding of the image. 
This paper incorporates local feature learning into self-supervised vision transformers via Reconstructive Pre-training (RePre).
Our RePre extends contrastive frameworks by adding a branch for reconstructing raw image pixels in parallel with the existing contrastive objective. 
RePre is equipped with a lightweight convolution-based decoder that fuses the multi-hierarchy features from the transformer encoder. 
The multi-hierarchy features provide rich supervisions from low to high semantic information, which are crucial for our RePre.
Our RePre brings decent improvements on various contrastive frameworks with different vision transformer architectures. 
Transfer performance in downstream tasks outperforms supervised pre-training and state-of-the-art (SOTA) self-supervised counterparts.
\end{abstract}

\begin{figure}
\centering
\includegraphics[width=0.5\textwidth]{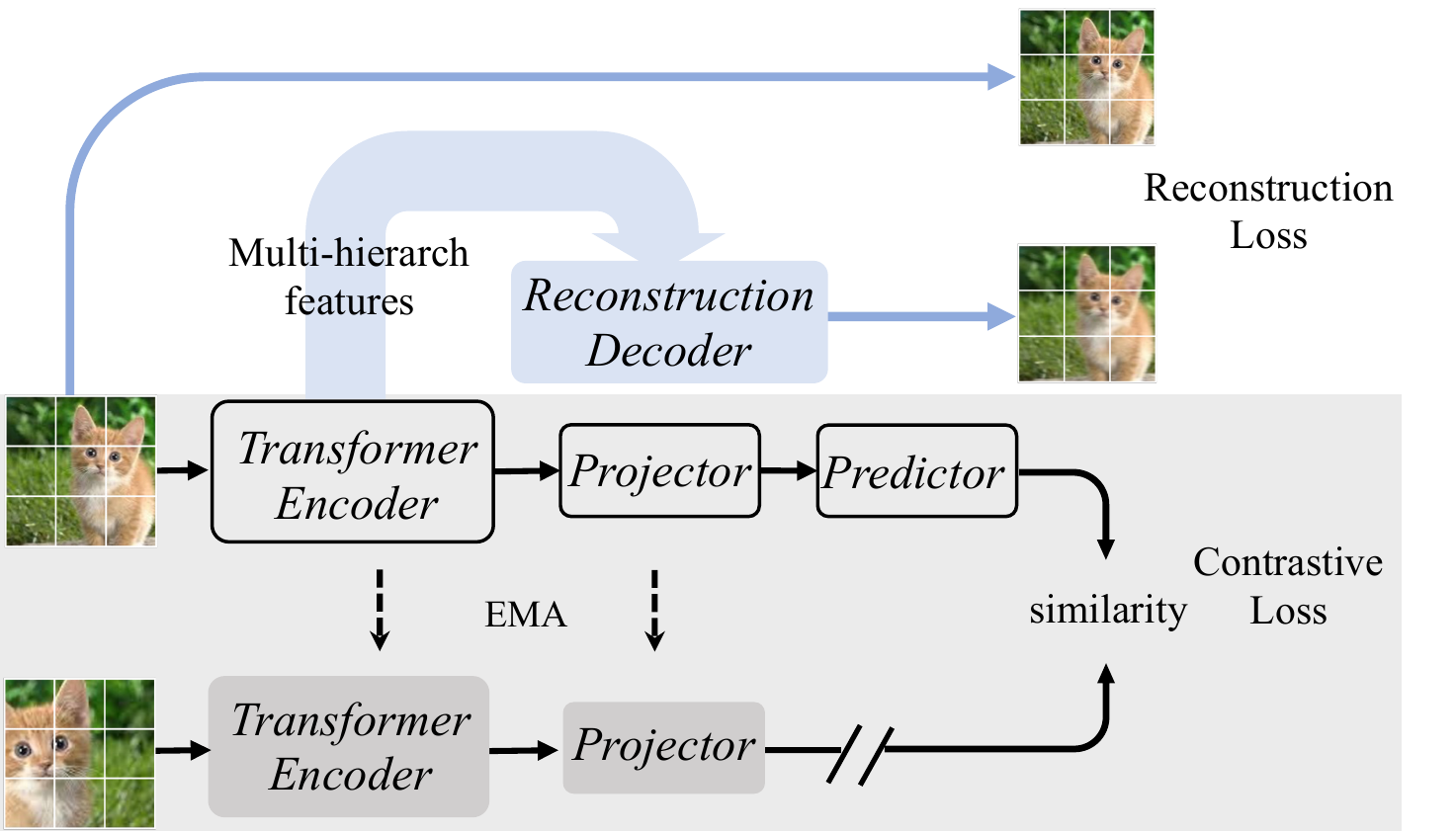} 

\caption{
Our RePre extends contrastive frameworks (bottom grey part) by adding a branch for reconstructing raw image pixels (top part). 
Contrastive framework (MoCo v3~\protect\cite{Alpher33} in this figure) models image similarity and dissimilarity between two views in an embedding space. 
Our reconstruction decoder recovers \emph{raw image pixels} using multi-hierarchy features from the transformer encoder.
}
\label{fig:repre_overview}
\end{figure}

\section{Introduction}
Self-supervised pre-training, a method to learn general representations without expensive annotated data, has greatly facilitated Natural Language Processing (NLP)~\cite{Alpher61,Alpher01,Alpher62} and similar trends also in Computer Vision (CV)~\cite{Alpher05,Alpher06,li2021supervision}.
One of the main ingredients for the success of self-supervised pre-training in NLP is using the scalable Transformer~\cite{Alpher26}, a self-attention-based architecture.
In CV, Vision Transformer (ViT)~\cite{Alpher27} has emerged as an      alternative to Convolutional Neural Networks (CNN) since its creation.
Despite its remarkable performance, pre-training the vanilla ViT requires enormous labeled data (e.g., JFT-300M~\cite{sun2017revisiting} in ~\cite{Alpher27}) and extensive computing resources. 
To avoid expensive labeled data, this paper studies pre-training self-supervised vision transformers.

There is a major difference between the self-supervised pre-training paradigm in NLP and CV: language transformers are pre-trained with masked/autoregressive language modeling~\cite{Alpher01,Alpher61}, while for vision transformers, the dominant method is contrastive learning which is based on instance discrimination pretext task ~\cite{Alpher33,Alpher34,Alpher51}. 
Concretely, contrastive learning maximizes the similarity of representations obtained from different views of the same image, leading to a global visual understanding (see the bottom part of Fig.~\ref{fig:repre_overview}). 
However, the sole global feature is insufficient for downstream tasks beyond image classification, such as object detection and segmentation~\cite{wang2021dense,xie2021detco}. 
Motivated by the intuition that a good visual representation should contain global features as well as fine-grained local features, we try to answer: \emph{could we achieve the best of both worlds?}

To achieve holistic visual representations, this paper incorporates fine-grained local feature learning in contrastive self-supervised vision transformers. 
Inspired by the widely used reconstructive pre-training in CV~\cite{bao2021beit}, NLP~\cite{Alpher01}, and speech~\cite{hsu2021hubert}, we choose a simple yet effective pretext task: \textbf{Re}construction \textbf{Pre}-training from raw pixels.
Intuitively, pixel reconstructing could let the network capture low semantics to learn fine-grained local features~\cite{ahn2018learning}.
Our RePre extends contrastive frameworks by adding a branch for reconstructing raw image pixels in parallel with the existing contrastive objective (see Fig.~\ref{fig:repre_overview}).
We split an image into patches and \emph{all} these RGB patches are reconstructed through a decoder. 
Worth mentioning, our neat RePre does not require masking strategy~\cite{hsu2021hubert,Alpher01} nor the tokenizer in BEIT~\cite{bao2021beit}.

Our initial trial is feeding the output of the last transformer encoder layer into the reconstruction decoder. 
However, it turns out this simple combination only brings marginal improvements. 
We argue that this ineffectiveness lies in the discrepancy between the last layer's high semantic features and the low semantic pixel objective.
Deep neural networks learn hierarchical semantic features via stacking layers~\cite{krizhevsky2012imagenet,he2016deep,Alpher27,Alpher50}. 
As the processing hierarchy goes up, the early layer captures simple low-level visual information (shallow features), and the late layer can effectively focus on complex high-level visual semantics (deep features).
Driven by this analysis, we propose to use the multi-hierarchy features in the transformer encoder. 
We collect the low to high semantic features within the transformer encoder and use them as a whole to guide the reconstruction.

\begin{figure*}[t]
\centering
\includegraphics[width=1\textwidth]{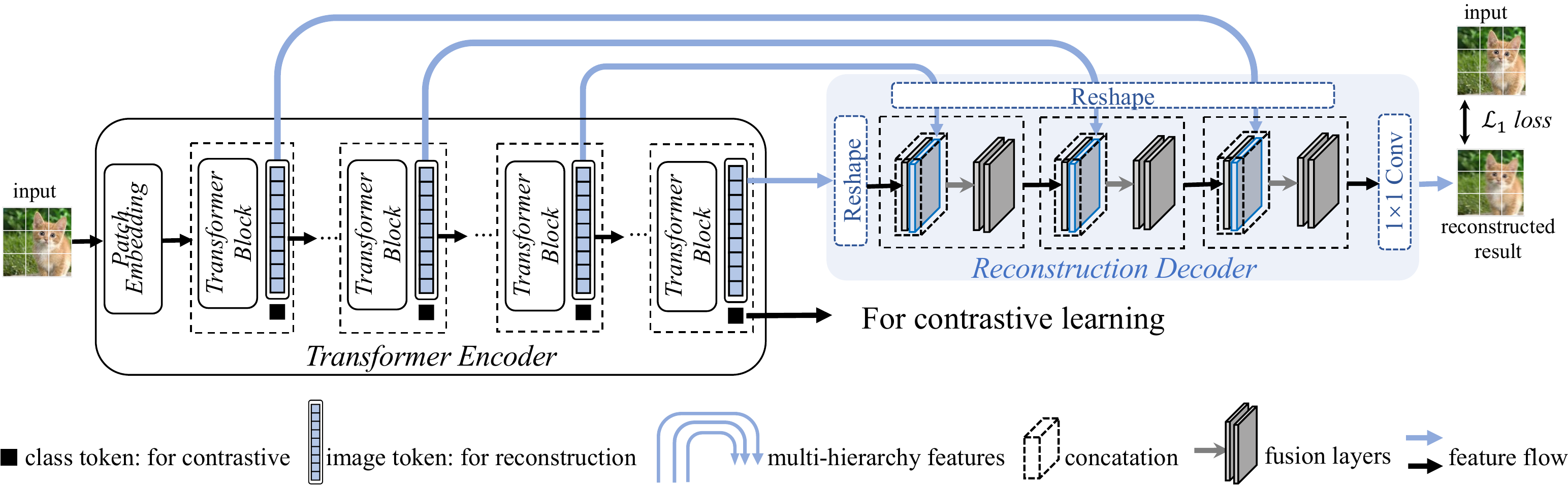} 
\caption{
Details of our reconstruction branch.
We get low-to-high semantic (multi-hierarchy) features from the transformer encoder by sampling shallow-to-deep transformer blocks. Our decoder gradually integrates deep-to-shallow features and regresses to predict the original RGB pixels with a simple $L_1$ loss.
The sequential image tokens are reshaped to 2D shape for convolution operators in reconstruction decoder.
The fusion block in decoder is simple: a concatenation followed by fusion layers.
Worth mentioning, for transformer variants with scale downsampling, such as the Swin Transformer, we need to upsample the high-level features before concatenation (details see Sec.~\ref{sec:reconstruction_decoder}).
}

\label{fig:repre_detail}
\end{figure*}

The reconstruction decoder is another essential part of our RePre.
Inspired by U-Net shape~\cite{Alpher52}, our decoder gradually integrates the deep to shallow features from multiple hierarchies and regresses to predict the original RGB pixels directly with a simple $L_{1}$ loss(see Fig.~\ref{fig:repre_detail}).
To combine multi-hierarchy features, the reconstruction decoder is consisted of several fusion layers.
Interestingly, we find that the fusion layer could be very \emph{lightweight}, e.g., one or two convolution layers.
Since our goal is to introduce additional local features while keeping the high-level semantic features intact, the heavy reconstruction decoder would focus too much on low semantic information, thus harming representation learning.
Another favorable property of a lightweight decoder is its little training overload.
Our RePre only brings a negligible average of 4\% workload in various contrastive frameworks.
The reconstruction decoder is only used during pre-training and dropped in the downstream fine-tuning phase.

Our RePre is generic and can be plugged into arbitrary contrastive learning frameworks for various visual translator architectures. Extensive experiments demonstrate the effectiveness and portability of this method. We validate our RePre in the latest contrastive learning frameworks (e.g., DINO, MOCO V3, MoBY, BYOL and SimCLR). Following standard linear evaluation on ImageNet-1K, with RePre, these methods improve top-1 accuracy by 0.5$\sim$1.1\%.
Prominently, it also brings significant performance to the base methods on dense prediction tasks on the COCO and cityscape datasets, even outperforming supervised methods.

Overall, our contributions are threefold: 
\begin{enumerate}
\item We incorporate fine-grained local feature learning in contrastive self-supervised vision transformers via adding a reconstruction branch.
We adopt a simple yet effective objective: Reconstructive Pre-training (RePre) from raw RGB pixels.

\item RePre utilizes multi-hierarchy fusion to provide rich supervisions from intermediate features.
We also find a fast lightweight convolutional reconstruction decoder could bring favorable results.

\item Our RePre is general and easy to be plugged. Decent improvements are observed on various contrastive frameworks with vision transformer and its variants. On dense prediction transfer tasks, RePre also brings significant improvements even outperforming supervised methods.
\end{enumerate}

\section{Related Work}

\subsection{Self-supervised vision transformer}

Self-supervised contrastive learning has been popular in computer vision. Before the emergence ViT, prior work mainly focus on ResNet~\cite{he2016deep}
, e.g, MoCo~\cite{Alpher19}, SimCLR~\cite{Alpher05}, BYOL~\cite{Alpher06}, SimSiam~\cite{Alpher25}. 
More recently, researchers have incorporated contrastive learning with ViT. 
MoCo v3~\cite{Alpher33} proposes an empirical study by training ViT with the MoCo framework. 
DINO~\cite{Alpher34} shows two new properties of self-supervised ViT compared with supervised ViT. 
MoBY~\cite{Alpher51} extends the contrastive framework with a ViT variant, Swin Transformer~\cite{Alpher50}. 
All these methods share the same spirit: modeling image similarity and dissimilarity (or only similarity) between two or more views, leading to a global image understanding.
They lack attention to local and low semantic features, which are crucial for downstream tasks beyond image classification, such as object detection and segmentation. 
Our RePre is complementary to these contrastive methods via enhancing fine-grained local feature learning.

\subsection{Reconstructive pre-training}

Reconstructive (or generative) objectives are highly successful for pre-training in NLP, e.g., masked/autoregressive language modeling in BERT~\cite{Alpher01} and GPT~\cite{Alpher61}. 
These methods hold out a portion of the input tokens and train models to predict the missing content.
In the field of CV, pioneering iGPT~\cite{Alpher18} learns a giant self-supervised transformer by directly predicting pixel values, producing competitive results with supervised counterparts. 
More recently, BEiT~\cite{bao2021beit} quantizes image patches as discrete tokens using an off-the-shelf discrete VAE(dVAE) tokenizer~\cite{dvae}, then proposes to predict the masked tokens. 
Following BEiT, iBoT~\cite{zhou2021ibot} introduces an online tokenizer.
Concurrent MAE~\cite{he2021masked} and SimMIM~\cite{xie2021simmim} propose to reconstruct raw pixels via mask image modeling. 
Differently, our RePre incorporates reconstructive pixel objectives along with contrastive learning frameworks. It pre-trains general vision transformers for various downstream tasks.
Moreover, our neat RePre reconstructs all the image pixels, so it does not require a masking strategy or tokenizer.

\section{Method}

In this section, we first discuss the contrastive learning frameworks. Then we introduce two key components in our RePre: multi-hierarchy features and a lightweight convolutional decoder (Fig.~\ref{fig:repre_detail}). Finally, we introduce the overall loss function of RePre.

\subsection{Revisiting contrastive learning frameworks}
\label{sec:contrastive_learning}

The primary focus of contrastive learning is to learn image embeddings that are invariant to different augmented views of the same image while being discriminative among different images. 
This is typically achieved by maximizing the similarity of representations obtained from different distorted versions of a sample using a variant of Siamese networks. 
As shown in the bottom part of Fig.~\ref{fig:repre_overview}: a Siamese network is composed of two branches: an online branch and a target branch, where the target branch keeps an Exponential Moving Average (EMA) of the online branch~\cite{Alpher33,Alpher51,Alpher34} or shares weights with the online branch~\cite{Alpher05,Alpher25} (not shown in Fig.~\ref{fig:repre_overview}). 
In particular, each branch encodes an augmented view to a single feature vector in the embedding space, resulting in a level of a global feature.

In order to better prove the scalability and effectiveness of our RePre in arbitrary contrastive learning frameworks, we roughly split current contrastive frameworks into two types: methods with negative samples, e.g., MoCo v3, SimCLR and methods without negative samples, e.g., BYOL, SimSiam. 

\textbf{Methods with negative samples} contrast positive samples with negative samples to prevent trivial solutions, i.e., all the outputs collapsing into constant. 
Specifically, augmented views created from the same samples are considered positive pairs, and images from different samples are considered negative pairs. 
The target branch outputs the representation of a positive sample and a set of negative samples, and the loss explicitly pulls the pair of positive samples together while pushing apart the pair of negative samples. 
The loss function can be thought of as a $K+1$ way softmax:
\begin{equation}
\mathcal L_{contrast\_w\_neg} = -log\frac{exp(q\cdot k_{+}/ \tau)}{\sum_{i=0}^{K} exp(q\cdot k_{i}/ \tau)}
\end{equation}
Where $k_{+}$ is the target feature for the other view of the same image; $k_{i}$ is a target feature of negative samples; $\tau$ is a temperature term; $K$ is the size of the queue or batch.

\textbf{Methods without negative samples} only rely on positive samples. 
They introduce asymmetric architecture to prevent collapse.  In particular, it appends a multi-layer perception as predictor to the encoder of the online branch, and it stops the gradient through the target branch. 
In this case, the loss explicitly pulls together the positive sample pairs, and the objective function is the negative cosine similarity between the two augmented views. 
Given the output of the online predictor $p_1$ and the output of the target branch $z_2$, the objective function is: 
\begin{equation}
\mathcal L_{contrast\_w/o\_neg} = -<\frac{p_1}{	\left \| p_1 \right \|_2},\frac{z_2}{	\left \| z_2 \right \|_2}>
\end{equation}
Where $<\cdot,\cdot>$ denotes the inner product operator.

\subsection{Reconstruction with multi-hierarchy features}
\label{sec:multi-hierarchy-feat}

Following the practice of ViT, we split the $H\times W\times 3$ shape image with patch size $P$. After patch embedding and linear projection, we get $\mathbf{z}_{0} \in \mathbb{R}^{(N+1)\times C}$, the sequential feature of an image, where $N=\frac{H}{P}*\frac{W}{P}$. The additional $1$ denotes the \texttt{class} token, $C$ is the number of channels. The sequential feature would iterate all $L$ transformer blocks within the encoder. We denote the output tokens of each block as $\{ \mathbf{z}_{1},\mathbf{z}_{2},...,\mathbf{z}_{L} \}$. In contrastive learning, $\mathbf{z}^0_L$ serves as the global image representation. For simplicity, we denote $\mathbf{z}$ excluding $\mathbf{z}^0$ as $\mathbf{y}$, which stands for the representations of patches. 
For Swin Transformer, because there is no \texttt{class} token, we get $\mathbf{y}_{0} \in \mathbb{R}^{N\times C}$ after patch embedding. 
Swin Transformer also has patch merging layers, which reduce the number of token by $\frac{1}{2}$ and increase the feature dimension by 2$\times$. The output embeddings of the last stage are averaged by a global average pooling layer and then sent to a linear classifier for classification, which is different from the class token used by ViT.

Our initial trial is feeding the output of the last transformer block $\mathbf{y}_{L}$ into the reconstruction decoder. However, it turns out this simple combination only brings marginal improvements (see Sec.~\ref{ablation:multi-hierarchy-feat}). We argue that this ineffectiveness lies in the discrepancy between the last layer’s high semantic features and the low semantic pixel objective. Inspired by U-Net shape, we collect low-to-high semantic features from shallow-to-deep blocks and reconstruct raw pixels gradually. Given a vanilla ViT with $L$ transformer blocks, we sample $K$  ($K<L$) hierarchical features with even intervals, i.e., our multi-hierarchy feature $\mathbf{Y} = \{ \mathbf{y}_{\floor{\frac{L}{K}}-1},\mathbf{y}_{\floor{\frac{L}{K}}*2-1},...,\mathbf{y}_{L-1} \}$, where $\floor{\cdot}$ is the floor function. $K=4$ is a standard practice in this paper. For a $L=12$ ViT-S, we sample $\mathbf{Y} = \{ \mathbf{y}_{2}, \mathbf{y}_{5}, \mathbf{y}_{8}, \mathbf{y}_{11} \}$ as multi-hierarchy feature.
For Swin Transformer, which has downsampling operators, we can also get the multi-hierarchy feature.
We directly sample the last feature of each resolution stage.

\subsection{Lightweight reconstruction decoder}
\label{sec:reconstruction_decoder}

With the multi-hierarchy feature, our decoder gradually integrates the deep-to-shallow features and regresses to predict the original RGB pixels directly with a simple $L_{1}$ loss(see Fig.~\ref{fig:repre_detail}). 
Surprisingly, we find that a lightweight convolutional decoder works pretty well (see Sec.~\ref{ablation:conv_decoder}), e.g., one or two fusion layers in each decoder block. A fusion layer consisits a $3\times3$ convolution layer and a ReLU layer.
In order to cooperate with the convolutional operator, the sequential feature $\mathbf{y} \in \mathbb{R}^{N\times C}$ is reshaped to 2D feature $ \mathbf{x} \in \mathbb{R}^{\frac{H}{P} \times \frac{W}{P}\times C}$. 
Like in U-Net, the shallow feature is merged into deep feature by concatenation, resulting in a feature with shape $\frac{H}{P} \times \frac{W}{P}\times 2C$.
In order to fuse the multi-hierarchy feature, our reconstruction decoder consists of fusion layers in each $K-1$ block (details in Fig.~\ref{fig:repre_detail}).
To predict all pixel values at a full resolution of input images, we apply a $1\times1$ convolution layer to map each feature vector in the final output of the decoder back to the original resolution. We let this vector take charge of the prediction of corresponding raw pixels.
Then, we apply a simple $L_{1}$ loss between the original image and the decoder output. In summary, the reconstruction objective is:
\begin{equation}
\mathcal L_{reconstruct} = |img - decoder(\mathbf{Y})| 
\end{equation}
Where $| \cdot |$ is the$L_{1}$ loss, $img$ is the augmented view before normalization, $\mathbf{Y}$ is the multi-hierarchy feature, $decoder(\cdot)$ returns the reconstructed image. 

Our decoder is also compatible with hierarchical vision transformers as Swin Transformer. Because of downsampling, we cannot directly concatenate the deep low-resolution feature with the shallow high-resolution feature. Thus, we apply a bilinear interpolation upsampling operation on the deep feature to make the alignment. 

\begin{algorithm}[tb]
\caption{Pseudo code of RePre in a PyTorch-like style}

\label{alg:algorithm}
\small
\begin{algorithmic}[1] 
\STATE \textcolor{codeblue}{\# reconst\_dec: convolutional reconstruction decoder}
\STATE \textcolor{codeblue}{\# online\_enc, target\_enc: transformer-based encoder}
\STATE \textcolor{codeblue}{\# online\_net = online\_enc + projector + predictor, predictor is None for symmetric methods}
\STATE \textcolor{codeblue}{\# target\_net = target\_enc + projector} 

\FOR{ x in loader:} 
\STATE v1, v2 = aug(x), aug(x)  \textcolor{codeblue}{\# augmentation} \\
\STATE \textcolor{codeblue}{ \# \# \# Reconstructive pre-training \# \# \#}\\
   multi\_hierarchy\_feats = online\_encoder(v1) \textcolor{codeblue} \\
\STATE reconst\_v = reconst\_dec(multi\_hierarchy\_feats) \textcolor{codeblue}{ \# reconst\_v with shape $H*W*3$}\\
\STATE    reconstruction\_loss = \textcolor{codekw}{$\mathcal{L}_{1}\_loss$}(reconst\_v, v1) \textcolor{codeblue}{ \# Details in Sec.~\ref{sec:reconstruction_decoder}}\\
\STATE   \textcolor{codeblue}{ \# \# \# Contrastive pre-training \# \# \#} \\
  q1, q2 = online\_net(v1), online\_net(v2) \textcolor{codeblue}{ \# queries: [N,C] each}\\ 
\STATE    k1, k2 = target\_net(v1), target\_net(v2) \textcolor{codeblue}{ \# keys: [N,C] each} \\
\STATE    contrast\_loss = \textcolor{codekw}{$ctr\_loss$}(\textit{q1}, \textit{k2}) + \textcolor{codekw}{$ctr\_loss$}(\textit{q2}, \textit{k1}) \textcolor{codeblue}{ \# Details in Sec.~\ref{sec:contrastive_learning}} \\
\STATE    \textcolor{codeblue}{ \# \# \# Combining objectives \# \# \#}\\
     loss = $\lambda_{1}$contrast\_loss + $\lambda_{2}$reconstruction\_loss
\STATE    loss.backward() 
\STATE    update(online\_net, reconst\_dec)\\ \textcolor{codeblue}{ \# target\_net is optimized by EMA or gradient}

\ENDFOR
\end{algorithmic}
\end{algorithm}

\subsection{Overall loss of RePre}

Our RePre is optimized with contrastive loss and reconstructive loss, which simultaneously learns the global features and fine-grained local features. The contrastive loss function is consistent with the contrastive learning method we use (details in Sec.~\ref{sec:contrastive_learning}). The reconstruction loss function computes the mean absolute error between the reconstructed and original images in pixel space (details in Sec.~\ref{sec:reconstruction_decoder}). 
We use a weighted sum of these two loss functions as our overall loss.
To avoid expensive calculation by optimizing the weights through the grid search method, we incorporate the uncertainty weighting approach proposed by~\cite{Alpher60}.
In particular, each task is weighted by a function of its homoscedastic aleatoric uncertainty rather than by a fixed weight. The overall loss function is then calculated as follows:
\begin{equation}
\mathcal L = \lambda_{1}\mathcal L_{contrast} + \lambda_{2}\mathcal L_{reconstruct}
\end{equation}
Where $\lambda_{1}$,$\lambda_{2}$ are learnable parameters.

\section{Experiments}

Our RePre is general and can be plugged into arbitrary contrastive learning frameworks with various vision transformer architectures. We first study the linear evaluation of the image recognition task. Then we transfer our pre-trained models into downstream object detection and semantic segmentation tasks. Finally, we do detailed ablation studies about the key components of our RePre.

\subsection{Linear Evaluation}
Linear evaluation on ImageNet-1K dataset is a standard evaluation protocol to assess the quality of learned representations. 
After pre-training, we add a linear layer on the top of the network. We only train this linear layer while fixing the pre-trained network.

\begin{figure}
\centering
\includegraphics[width=0.5\textwidth]{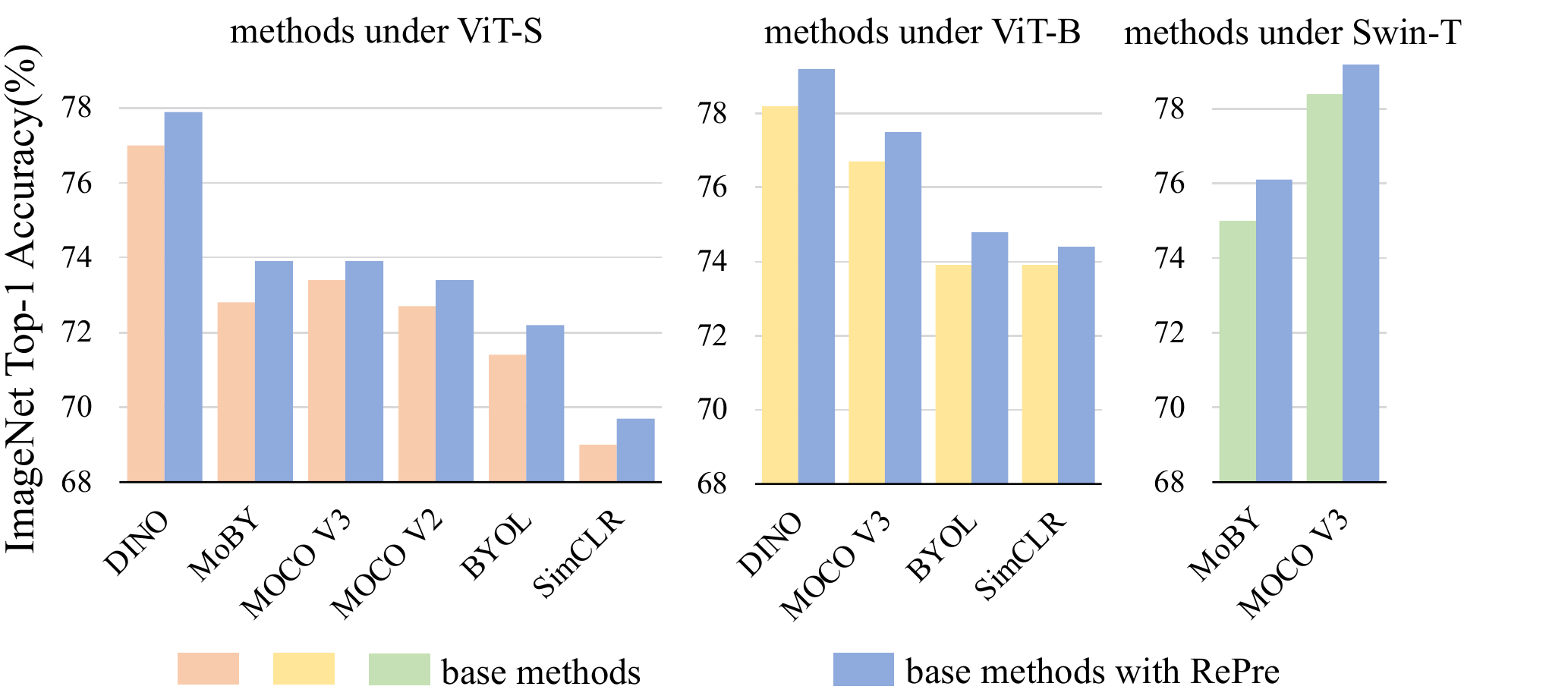} 
\caption{Performance improvements brought by RePre when using different contrastive learning frameworks and network architectures. }
\label{fig:performance_lp}
\end{figure}

Fig.~\ref{fig:performance_lp} and Table~\ref{table:top1-acc} list the apparent performance improvements that our RePre brings to different advanced comparative learning methods based on different backbone architectures.
Our pre-training and fine-tuning recipes are basically the same as the contrastive learning methods. Since our reconstruction decoder is lightweight, our RePre only brings a negligible average of 4\% workload. All our experiments are conducted on NVIDIA V100 GPUs.

\begin{table}
\centering
\begin{tabular}{lllll}
\toprule
Base&Arch.&Epoch&Acc\%&Acc w/ RePre\\
\midrule
    %
    SimCLR&&300& 69.0& \textbf{69.7}($\uparrow$ 0.7)\\
    %
    BYOL&&300& 71.4& \textbf{72.2}($\uparrow$ 0.8)\\
    %
    MOCO v2&&300& 71.6& \textbf{72.1}($\uparrow$ 0.5)\\
    MOCO v2&&800& 72.7& \textbf{73.4}($\uparrow$ 0.7)\\
    %
    MOCO v3&ViT-S&300& 72.5& \textbf{73.2}($\uparrow$ 0.7)\\
    MOCO v3&&600& 73.4& \textbf{73.9}($\uparrow$ 0.5)\\
    %
    %
    MoBY&&300& 72.8& \textbf{73.9}($\uparrow$ 1.1)\\
    DINO &&300& 75.9& \textbf{76.7}($\uparrow$ 0.8)\\
    DINO&&800& 77.0& \textbf{77.9}($\uparrow$ 0.9)\\

    \midrule
    %
    SimCLR&&300& 73.9& \textbf{74.4}($\uparrow$ 0.5)\\
    %
    %
    BYOL&&300& 73.9& \textbf{74.8}($\uparrow$ 0.9)\\
    %
    MOCO V3&ViT-B&300& 76.5&\textbf{77.2}($\uparrow$ 0.7)\\
    MOCO V3&&600&  76.7 &\textbf{77.5}($\uparrow$ 0.8)\\
    DINO&&800& 78.2& \textbf{79.2}($\uparrow$ 1.1)\\
    %
    \midrule
    %
    MOCO v3&&300& 75.4$^*$& \textbf{76.4}($\uparrow$ 1.0)\\
    %
    MoBY&Swin-T&100& 70.9& \textbf{71.8}($\uparrow$ 0.9)\\
    MoBY&&300& 75.0& \textbf{76.1}($\uparrow$ 1.1)\\
\bottomrule
\end{tabular}
\caption{More linear evaluation results on ImageNet benchmark. Our RePre brings consistent gains under different methods, architectures and training epochs. $^*$ is our reimplementation.}
\label{table:top1-acc}
\end{table}

\subsection{Transfer to downstream tasks}
We further evaluate the transferring performance of the learned representations on downstream tasks of COCO~\cite{Alpher54} object detection/instance segmentation and Cityscapes~\cite{Alpher59} semantic segmentation.

\subsubsection{Object detection and instance segmentation}
We perform object detection/instance segmentation experiments on COCO with Mask R-CNN~\cite{Alpher31}
framework. Following standard practice, we use AdamW optimizer and $1\times$ schedule. The shorter edges of the input images are resized to 800 while the longer side is at most 1333.
To compare with advanced research results, we use Swin-T as the backbone. As shown in Table~\ref{table:coco-detect}, the performance of MoBY with RePre is improved by 1.2\% and 0.7\% under the same pre-training settings. Similarly, our RePre brings effective performance gains of 0.9\% and 1.3\% for DINO.

\begin{table}
\centering
\begin{tabular}{lllll}
\toprule
 Method&w/ RePre&Epoch&mAP$^{bbox}$&mAP$^{mask}$\\
\midrule
    IN Sup.      &$-$   &100&	41.6&		38.4\\
          &$-$   &300&	43.7&   	39.8\\
    \midrule
          &   \texttimes&100&	41.5&       38.3\\
    MoBY&\checkmark&		100&	\textbf{42.1}($\uparrow$ 0.6)&\textbf{39.2}($\uparrow$ 0.8)\\
          &   \texttimes&300&   43.6&       39.6\\
    &\checkmark&		300&	\textbf{44.8}($\uparrow$ 1.2)&\textbf{40.3}($\uparrow$ 0.7)\\
    \midrule
    DINO      &    \texttimes&100&	42.2&       38.7\\
    &\checkmark&       100&
    \textbf{43.1}($\uparrow$ 0.9)&  \textbf{40.0}($\uparrow$ 1.3)\\
\bottomrule
\end{tabular}
\caption{Results of object detection and instance segmentation fine-tuned on MS COCO. We use Mask R-CNN framework with Swin-T as the backbone. Our RePre models outperform supervised ImageNet pre-training and self-supervised DINO with a decent margin.}
\label{table:coco-detect}
\end{table}

\begin{table}
\centering
\begin{tabular}{cccc}
\toprule
 Method&	Arch. &mIoU&	mAcc(\%)\\
\midrule
    IN Supervised&	&71.33&	80.30\\
    DINO& ViT-S	&72.96&	81.32\\
    \textbf{DINO w/ RePre}&	&\textbf{73.40}&	\textbf{81.95}\\
\bottomrule
\end{tabular}
\caption{Results of semantic segmentation fine-tuned on Cityscapes. We use SETR framework with ViT-S as the backbone. All the ViT-S modelsare all pre-trained for 300 epochs.}
\label{table:seg}
\end{table}

\subsubsection{Semantic segmentation}
We adopt the SETR~\cite{Alpher32} as the semantic segmentation strategy on Cityscapes and follow the training config as original SETR. For a fair comparison, we use pretrained models based on ViT-S by 300-epoch. As shown in Table~\ref{table:seg}, DINO with RePre achieves the highest mIoU 73.40\% and mAcc 81.95\%. It outperforms both supervised and DINO pretrained results. It proves that reconstruction pretraining extracts finer local-level information and is suitable to transfer for semantic segmentation tasks.

\subsection{Ablation Study}
Two key components of our RePre are multi-hierarchy features and reconstruction decoder. In this part, we perform detailed ablation studies on these two components. Without specification, we use MOCO v3 as the contrastive learning framework and the pre-training epoch is 300.

\begin{table}
\centering
\begin{tabular}{ccccc}
\toprule
Method &Arch.& Single& Multi&Top-1 Acc(\%)\\
\midrule
    MOCO v3&&--&--	&72.5\\
    MOCO v3&&\checkmark&	&72.8\\
    MOCO v3&ViT-S&&\checkmark	&\textbf{73.2}\\
    MoBY&&--&--&72.8\\
    MoBY&&\checkmark&	&73.1\\
    MoBY&&&\checkmark	&\textbf{73.9}\\
    \midrule
    MOCO v3&&--&--	&75.4\\
    MOCO v3&&\checkmark&	&75.7\\
    MOCO v3&Swin-T&&\checkmark	&\textbf{76.4}\\
    MoBY&&--&--&75.0 \\
    MoBY&&\checkmark&	&75.4\\
    MoBY&&&\checkmark	&\textbf{76.1}\\
\bottomrule
\end{tabular}
\caption{Ablation study on positive impact of multi-hierarchy features. `Single' and 'Multi' denotes using the last layer output features or using the fused multi-hierarchy features respectively.}
\label{table:Multi-hierarchy-features}
\end{table}

\subsubsection{Ablation on multi-hierarchy features}
\label{ablation:multi-hierarchy-feat}

Tab.~\ref{table:Multi-hierarchy-features} shows the impact of multi-hierarchy features on performance with the our default convolutional decoder. As we can see, using the `Single' feature (last layer's output) only brings marginal improvements. We argue that this ineffectiveness lies in the discrepancy between the last layer’s high semantic features and the low semantic pixel objective. Using multi-hierarchy features (dubbed `Multi'), RePre improves MOCO v3 top-1 accuracy performance by 0.7\% under DeiT-S and 1.0\% under Swin-T. RePre also improves MoBY baseline top-1 accuracy performance by 1.1\% under DeiT-S or Swin-T. We conjecture that multi-hierarchy features contains different level of semantic information which is crucial for the reconstruction.

\begin{figure}
\centering
\includegraphics[width=0.5\textwidth]{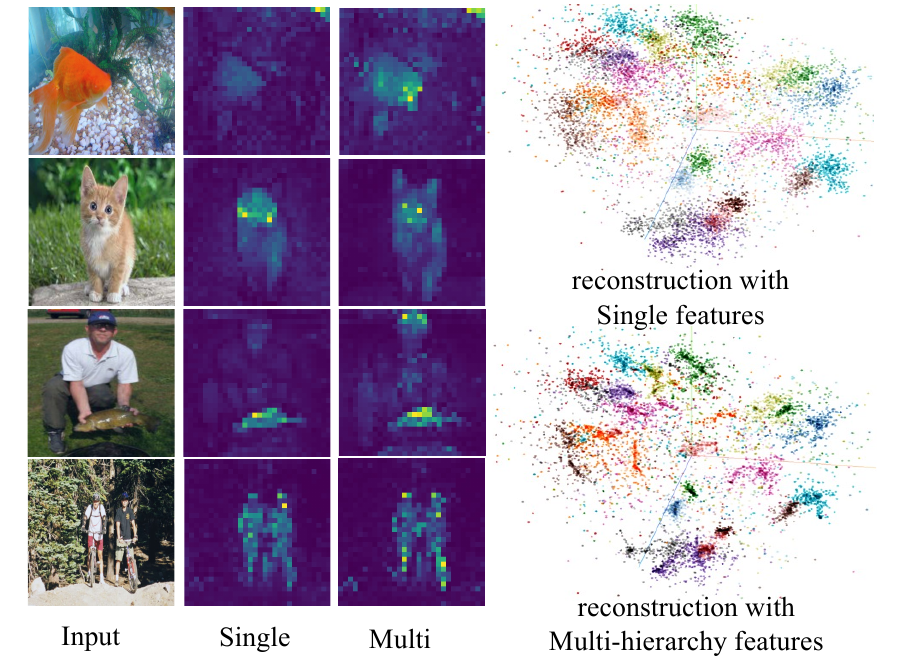} 
\caption{\textbf{Left part:} The visualization of attention map. The first column is original images. The second and the third column show \texttt{class} token's attentions when using last layer's or multi-hierarchy features.When using multi-hierarchy features, models can identify the edge area of the object more accurately and highlight the core focus.
\textbf{Right part:} The visualization of t-SNE on ImageNet. We randomly select 20 classes in the validation set. Each point represents embedding from online transformer encoder.}
\label{fig:ijcai_feature}
\end{figure}

We also show the comparison via showing the attention map and t-SNE in Fig.~\ref{fig:ijcai_feature}. As we can see from the left part, when using multi-hierarchy features, models can identify the edge area of the object more accurately and highlight the core focus. The phoneme can also explain the excellent performance of our RePre when transferring to detection and segmentation tasks. 
The right part in Fig.~\ref{fig:ijcai_feature} shows t-SNE~\cite{Alpher56} visualization results. Obviously, the learned representations with multi-hierarchy features can be better divided into different categories.

\begin{table}
\centering
\begin{tabular}{cccc}
\toprule
Operator & Layer Num.	&Arch.&	Top-1 Acc(\%)\\
\midrule
    w/o decoder&--&		&72.5\\
    Conv&1&	    &73.0\\
    Conv&2&&\textbf{73.2}\\
    Conv&4&ViT-S    & 73.2 \\
    Transformer&1 &		&71.8\\ 
    Transformer&2 &		&72.0\\ 
    Transformer&4 &		&71.4\\ 
    \midrule
    w/o decoder&--&	    &75.4\\
    Conv&1& 	&76.1\\
    Conv&2&&\textbf{76.4}\\
    Conv&4& Swin-T   & 76.2  \\
    Transformer&1 &		&74.6\\ 
    Transformer&2 &		&75.2\\ 
    Transformer&4 &		&74.5\\ 
\bottomrule
\end{tabular}
\caption{Ablation study on the fusion layer in reconstruction decoder. Operator and layer number denotes the type and the number of fusion layers.}
\label{table:decoder-arch}
\end{table}

\subsubsection{Ablation on fusion layer in reconstruction decoder}
\label{ablation:conv_decoder}
We analyze that convolution can enhance the fine-grained local spatial correlation without damaging context semantics information.
We validate it using the same basic Transformer layer as the backbone to replace the $3\times3$ convolution (Conv) in the decoder fusion layer. Table~\ref{table:decoder-arch} shows the positive effect brought by convolution. `Layer Num' represents the repetition times of fusion layer. With a lightweight convolutional decoder, RePre improves baseline top-1 accuracy performance by 0.7\% with ViT-S and 1.0\% with Swin-T, which could be a strong proof of our hypothesis. 
The results also validate our analysis that the heavy convoltion or transformer reconstruction decoder would focus too much on low semantic information, thus harming representation learning.

\section{Conclusion}
This work proposes a simple yet effective objective: Reconstructive Pre-training (RePre) from raw RGB pixels to train self-supervised vision transformers. 
Our RePre extends contrastive frameworks by adding a branch for reconstructing raw image pixels in parallel with the existing contrastive objective. 
RePre incorporates local feature learning with a lightweight convolution-based decoder that fuses the multi-hierarchy features from the transformer encoder.
Our RePre is generic and can improve arbitrary contrastive learning frameworks with negligible additional cost. 
With the fact that self-supervised learning in CV was dominated by contrastive objectives in the past several years, we hope our RePre could bring more insights into reconstructive (generative) objectives in this area.



\bibliographystyle{named}

\end{document}